\documentclass[nohyperref]{article}

\usepackage{microtype}
\usepackage{graphicx}
\usepackage{subfigure}
\usepackage{booktabs} 
\usepackage{hyperref}

\usepackage[accepted]{icml2022_TAGML}

\usepackage{amsmath}
\usepackage{amssymb}
\usepackage{mathtools}
\usepackage{amsthm}

\usepackage[capitalize,noabbrev]{cleveref}

\theoremstyle{plain}

\theoremstyle{definition}

\theoremstyle{remark}

\newcommand{\diag}{\textnormal{diag}}

\newcommand{\rank}{\textnormal{rank}}

\newcommand{\R}{
\mathbb{R}}

\def\T{\mathcal{T}}

\def\fro{\mathrm{F}}

\newcommand{\eps}{\varepsilon}

\usepackage[textsize=tiny]{todonotes}

\begin{document}

\twocolumn[
\icmltitle{Riemannian CUR Decompositions for Robust Principal Component Analysis}



\icmlsetsymbol{equal}{*}

\begin{icmlauthorlist}
\icmlauthor{Keaton Hamm}{UTA}
\icmlauthor{Mohamed Meskini}{UTA}
\icmlauthor{HanQin Cai}{UCLA}
\end{icmlauthorlist}

\icmlaffiliation{UTA}{Department of Mathematics, University of Texas at Arlington, Arlington, TX, USA}
\icmlaffiliation{UCLA}{Department of Mathematics, University of California, Los Angeles, Los Angeles, CA, USA}

\icmlcorrespondingauthor{HanQin Cai}{hqcai@math.ucla.edu}

\icmlkeywords{Robust PCA, Nonconvex Optimization, Low-rank Matrix, Riemannian Optimization, Manifold Method}

\vskip 0.3in
]



\printAffiliationsAndNotice{}  

\begin{abstract}
Robust Principal Component Analysis (PCA) has received massive attention in recent years. It aims to recover a low-rank matrix and a sparse matrix from their sum. This paper proposes a novel nonconvex Robust PCA algorithm, coined Riemannian CUR (RieCUR), which utilizes the ideas of Riemannian optimization and robust CUR decompositions. This algorithm has the same computational complexity as Iterated Robust CUR, which is currently state-of-the-art, but is more robust to outliers. RieCUR is also able to tolerate a significant amount of outliers, and is comparable to Accelerated Alternating Projections, which has high outlier tolerance but worse computational complexity than the proposed method. Thus, the proposed algorithm achieves state-of-the-art performance on Robust PCA both in terms of computational complexity and outlier tolerance.


\end{abstract}




\section{Introduction}\label{SEC:Introduction}

Principal Component Analysis (PCA) is one of the most fundamental tools to uncover low-dimensional structure in high-dimensional data. Robust PCA proposed by \citep{wright2009robust,candes2011robust} seeks to find an incoherent, low-rank matrix from observations of it corrupted by sparse outliers \citep{bouwmans2018applications}.  That is, given observations of the form $D = L+S$, where $L$ is the underlying low-rank signal and $S$ is a matrix of sparse, but arbitrary magnitude outliers, one seeks to find a good approximation of $L$. Robust PCA has a wide range of applications, for instance, video static background subtraction \citep{jang2016primary}, singing-voice separation \citep{huang2012singing}, ultrasound imaging \citep{cai2021lrpca}, face recognition \citep{wright2008robust}, image alignment \citep{peng2012rasl}, community detection \citep{chen2012clustering}, and NMR spectrum recovery \citep{ASAP_Hankel,cai2022structured}.

To get around the computational cost of the semidefinite program for solving the convex relaxation to Robust PCA, many recent methods attack the nonconvex optimization problem directly:
\begin{equation}
\begin{aligned}
& \underset{L,S}{\text{minimize}}
& & \|D-L-S\|_\fro \\
& \text{subject to}
& & \rank(L)\leq r, \\
&&& \|S\|_0\leq\Delta.
\end{aligned}
\end{equation}
Several recent algorithms use an iterative procedure in which one forms alternating approximations of $L$ and $S$, in which $L$ is updated by projecting $D-S$ onto the Riemannian manifold of rank-$r$ matrices \citep{vandereycken2013low}, denoted $\mathcal{M}_r$, and $S$ is updated by projecting $D-L$ onto the set of sparse matrices that have at most $\Delta$ non-zero entries, denoted $\mathcal{S}_\Delta$. This technique is called Alternating Projections (AltProj) \citep{netrapalli2014non}.  One drawback of this algorithm is that it requires computation of the SVD of the full $n\times n$ matrix $D-S$, which carries complexity $\mathcal{O}(n^2r)$.  \citep{cai2019accelerated} propose Accelerated Alternating Projections (AccAltProj) which instead projects $D-S$ onto the tangent space of manifold $\mathcal{M}_r$ at the current estimation of $L$, which is significantly faster, while remaining robust to outliers. 
The trick of tangent space projection in AccAltProj coincides with the idea of Riemannian optimization \citep{wei2020guarantees}, thus AccAltProj can be viewed as a manifold method. 

Subsequently, \citep{cai2020rapid} proposed the use of CUR decompositions \citep{drineas2006fast,mahoney2009cur,hamm2020perspectives} to provide an even faster nonconvex Robust PCA solver, Iterated Robust CUR (IRCUR). IRCUR reduced the computational complexity of AccAltProj and previous methods from $\mathcal{O}(n^2r)$ to $\mathcal{O}(r^2n\log^2n).$ However, one drawback of IRCUR is that it is not capable of handling as many outliers as the previous methods (i.e., $\Delta$ must be smaller for IRCUR to be successful compared to AccAltProj).  To bridge this gap, this paper proposes \textit{Riemannian CUR} (RieCUR), which combines the idea of using CUR decompositions with the tangent space projections of AccAltProj. 

The result is an algorithm that combines the best of both methods in the following way:
\begin{itemize}
    \item  RieCUR has complexity $\mathcal{O}(r^2n\log^2n)$, which is the same as IRCUR (although the constant appears to be larger for RieCUR based on experiments in the sequel).
    \item  RieCUR appears to tolerate outliers as well as AccActProj in terms of reconstruction vs. sparsity based on our numerical experiments, whereas IRCUR degrades as the amount of outliers increases.
\end{itemize}




Additionally, many of the nonconvex solvers initialize guesses of $L$ and $S$ via a single truncated SVD of $D$. We show in our experiments here that one can use fast matrix sketching approaches for approximating the SVD during initialization while not sacrificing recovery of $L$ and $S$.  This allows for significant speedups in practical settings.

\section{Proposed Approach}\label{SEC:Algorithm}

Here we detail our algorithm, coined RieCUR, as well as its implementation details.  First, let us collect some notation and assumptions. 

\subsection{Notation}

We focus on square matrices $D,L,S\in\R^{n\times n}$, but note the proposed algorithm works for rectangular matrices (see \citep{cai2019accelerated} for details on converting rectangular Robust PCA problems to square ones). The SVD of $A\in\R^{n\times n}$ is $W\Sigma V^\top$ (or $W_A\Sigma_AV_A^\top$ if it is unclear from context), where $W$ and $V$ are orthogonal (called the left and right singular vectors, respectively) and $\Sigma=\diag(\sigma_1,\dots,\sigma_n)$ with $\sigma_1\geq\dots\geq\sigma_n\geq0$ (called the singular values of $A$. The truncated SVD of $A$ with rank parameter $r$ is denoted $A_r=W_r\Sigma_rV_r^\top$, where $W_r$ and $V_r$ are the $n\times r$ submatrices of $W$ and $V$, respectively corresponding to choosing the first $r$ columns of each, and $\Sigma_r$ is the $r\times r$ leading principal minor of $\Sigma$. The Frobenius norm of a matrix is $\|A\|_\fro:=(\sum_{i,j}|A_{ij}|^2)^\frac12,$ the $\ell_0$-``norm" of a matrix is the number of nonzero entries of it, i.e., $\|S\|_0:=|\{(i,j):S_{ij}\neq0\}|$ (here $|\cdot|$ denotes set cardinality), and $\|\cdot\|_2$ will here always denote the Euclidean norm of a vector.

Given an index set $I\subset\{1,\dots,n\}$, $A_{I,:}$ denotes the $|I|\times n$ submatrix of $A$ corresponding to choosing the rows of $A$ indexed by $I$, and similar meanings are assigned to $A_{I,J}$ and $A_{:,J}$.  Finally, if $A = W\Sigma V^\top$, then its Moore-Penrose pseudoinverse \citep{penrose1956best} is $A^\dagger = V\Sigma^\dagger W^\top$, where $\Sigma^\dagger$ contains diagonal entries $1/\sigma_i$ if $\sigma_i>0$ or $0$ if $\sigma_i=0$ along its diagonal.

Finally, given $\zeta>0$, we define the hard thresholding operator by
\[[\T_{\zeta}A]_{i,j} := \begin{cases} A_{i,j}, & |A_{i,j}|\geq \zeta\\ 0, & \textnormal{otherwise.}\end{cases}\]

\subsection{Assumptions}

We utilize common assumptions for Robust PCA problems to be tractable, namely incoherence and sparsity.  A matrix $L\in\R^{n\times n}$ with truncated SVD $W_r\Sigma_rV_r^\top$ is called $\mu$-incoherent provided 
\[\max_{i}\|(W_r)_{i,:}\|_2\leq \sqrt{\frac{\mu r}{n}},\quad \max_j \|(V_r)_{j,:}\|_2 \leq \sqrt{\frac{\mu r}{n}}.\]

A matrix $S\in\R^{n\times n}$ is said to be $\alpha$-sparse provided
\[\|S_{i,:}\|_0\leq \alpha n, \quad \|S_{:,j}\|_0 \leq \alpha n,\quad \textnormal{ for all  } i,j=1,\dots,n.\]
Note that this notion of sparsity requires that $S$ not have an overly large concentration of outliers on any given row or column.

\subsection{Review of CUR decompositions and tangent space projections}

Here we recall the background of both CUR decompositions and tangent space projections which are used in our main algorithm.

The CUR decomposition of a matrix comes from the observation that if one chooses a column submatrix $C = A_{:,J}$ whose columns span the column space of $A$ and a row matrix $R = A_{I,:}$ whose rows span the row space of $A$, and let $U=A_{I,J}$, then $A=CU^\dagger R$. See \citep{goreinov1997theory,hamm2020perspectives,strang2022lu} for more details. Based on random sampling methods for CUR decompositions of incoherent low-rank matrices \citep{chiu2013sublinear}, we sample $\mathcal{O}(r\log n)$ columns and rows for each CUR decomposition done in this work.

During the iterative procedure of Riemannian CUR, if we have estimate $L_k$ and $S_k$ of $L$ and $S$ at iteration $k$, and $L_k$ has truncated SVD of order $r$ given by $W_k\Sigma_kV_k^\top$, then the singular vectors $U_k$ and $V_k$ form a $(2n-r)r$-dimensional subspace of $\mathcal{M}_r$, which is the tangent space at $L_k$, defined by
\[T_k:= \{W_kA^\top+BV_k^\top:A,B\in\R^{n\times r}\}.\]
One can verify \citep{vandereycken2013low} that the projection onto $T_k$ is given by
\begin{equation}\label{EQN:PTk}
P_{T_k}X = W_kW_k^\top X + XV_kV_k^\top - W_kW_k^\top XV_kV_k^\top.\end{equation}

The main idea of AccAltProj \citep{cai2019accelerated} is that $D-L_k$ is projected onto $T_k$ and then projected onto $\mathcal{M}_r$ to obtain $S_{k+1}$; then $L_{k+1}$ is obtained by hard thresholding $D-S_{k+1}$.  The reason this speeds up the procedure of AltProj is that one can compute the SVD of $P_{T_k}(D-S_k)$ much more efficiently than the SVD of $D-S_k$ itself.

The main idea of IRCUR \citep{cai2020rapid} and its variants \citep{cai2021robust,cai2021fast} is that one only every works with column or row submatrices of $D$, $L_k$, and $S_k$ in the iterative process. That is, one is only trying to recover $L_{I,:}$, $L_{:,J}$, $S_{I,:}$, and $S_{:,J}$ for given row and column subsets $I$ and $J$, and at the end, one approximates $L$ by a CUR decomposition of the form $L \approx CU^\dagger R$.  Since only a few columns and rows of $D$ are utilized and held in memory, this algorithm is significantly faster than previous ones that required operations on the entire $n\times n$ matrices in question.

\subsection{Riemannian CUR algorithm}

We are now ready to state our algorithm below. The stopping criterion is in terms of an empirical relative error defined by
\[e_k = \frac{\|(D-L_k-S_k)_{I.:}\|_\fro+\|(D-L_k-S_k)_{:,J}\|_\fro}{\|D_{I,:}\|_\fro+\|D_{:,J}\|_\fro},\] where $I$ and $J$ are the indices of the submatrices of $P_{T_k}(D-S_k)$ formed at each iteration. This is the same error term as in IRCUR, and is used because our algorithm does not form all of $L$ or $S$ during the iterative part, but rather only tracks submatrices of them.

\begin{algorithm}[h!]
 \caption{Riemannian CUR (RieCUR) for Robust PCA}\label{ALG:Main}
\begin{algorithmic}[1]

\STATE 
\textbf{Input: }$D$: observed corrupted data matrix; $r$: rank; $\eps$: target precision level; $\zeta_0$: initial threshold; $\gamma$: threshold decay rate; $|I|,|J|$: number of rows
and columns to sample.
\STATE 
Uniformly sample row indices $I$ and column indices $J$ 

\STATE \textbf{Initialize} $L_0$ and $S_0$

\STATE $k=0$

\WHILE{$e_k>\eps$}
\STATE (Optional:) Resample row and column indices

\STATE $C_{k+1} = (P_{T_k}(D-S_{k}))_{:,J}$

\STATE $R_{k+1} = (P_{T_k}(D-S_{k}))_{I,:}$

\STATE $U_{k+1} = (P_{T_k}(D-S_{k}))_{I,J}$

\STATE $L_{k+1} = C_{k+1}U_{k+1}^\dagger R_{k+1}$

\STATE $\zeta_{k+1} = \gamma^k\zeta_0$

\STATE $(S_{k+1})_{:,J} = \T_{\zeta_{k+1}}(D-L_{k+1})_{:,J}$

\STATE $(S_{k+1})_{I,:} = \T_{\zeta_{k+1}}(D-L_{k+1})_{I,:}$


\STATE $k=k+1$
 \ENDWHILE
 \STATE \textbf{Output: }$C_k$, $U_k$, $R_k$: CUR decomposition of $L$
\end{algorithmic}
\end{algorithm}

The main difference of Algorithm~\ref{ALG:Main} compared with AccAltProj and IRCUR is that it utilizes submatrices of the tangent space projection $P_{T_k}(D-S_k)$ rather than the entire matrix as AccAltProj does, and IRCUR does not utilize the tangent space projection at all, but rather works with submatrices of $D-S_k$ directly. Use of the tangent space projection in RieCUR adds a small amount of computation time to IRCUR but with the benefit of making the procedure more robust to outliers, as CUR decompositions are known to suffer from outliers.

\subsection{Implementation details}

There are three primary details to consider in the implementation of Algorithm~\ref{ALG:Main}: initialization, forming the column and row submatrices of $P_{T_k}(D-S_k)$, and how to choose the initial threshold $\gamma$, which here is a hyperparameter of the algorithm, but we note that one can use the choices of \citep{cai2019accelerated}, and the algorithm regularly converges in thorough experimentation; however, at present we leave this as a general parameter and leave the analysis of choosing an adaptive threshold value to future work.

For initialization, we use a modification of Algorithm 3 of \citep{cai2019accelerated}, which uses two steps of Alternating Projections \citep{netrapalli2014non} to obtain $L_0$ and $S_0.$  

\begin{algorithm}[h!]
\caption{Initialization via CUR Decomposition}\label{ALG:Init}
\begin{algorithmic}[1]

\STATE 
\textbf{Input: }$D$: observed corrupted data matrix; $r$: target rank; $\beta_1,\beta_2$: thresholding parameters; $|I|,|J|$: number of rows and columns to sample.

\STATE 
Uniformly sample row indices $I$ and column indices $J$ 

\STATE $S_{-1} = \T_{\beta_1}(D)$

\STATE $C = (D-S_{-1})_{:,J}$

\STATE $R = (D-S_{-1})_{I,:}$

\STATE $U = (D-S_{-1})_{I,J}$

\STATE $L_0 = CU_r^\dagger R$

\STATE $S_0 = \T_{\beta_2}(D-L_0)$

\STATE \textbf{Output:} $L_0, S_0$
\end{algorithmic}
\end{algorithm}

We do allow one significant change in our experimental examination of Algorithm~\ref{ALG:Main}: when we initialize via Algorithm~\ref{ALG:Init}, rather than computing a full SVD of $D-S_{-1}$ in line 4 as in \cite{cai2019accelerated}, we instead compute a CUR decomposition of it (lines 4-6). This change reduces the computational cost significantly. As noted above, we sample $\mathcal{O}(r\log n)$ columns and rows of $D-S_{-1}$ in forming the CUR decomposition. Thus, computing the pseudoinverse of the intersection matrix requires only $\mathcal{O}(r^3 \log^2n)$ flops as opposed to $\mathcal{O}(n^2r)$ to compute the SVD of $D-S_{-1}$.

Finally, one of the speed advantages of IRCUR is that one need only operate on small submatrices of $D-S_k$ at a time. This is not quite the case for Riemannian CUR, as one cannot form a column submatrix of $P_{T_k}(D-S_k)$ by only using the corresponding columns of $D-S_k$. Nonetheless, we are not required to form all of $P_{T_k}(D-S_k)$ as is done in AccAltProj here. Indeed, at time instance $k$, we have $D-S_k = C_kU_k^\dagger R_k$, so
\begin{multline*}P_{T_k}(D-S_k) = W_kW_k^\top C_kU_k^\dagger R_k + C_kU_k^\dagger R_kV_kV_k^\top \\ - W_kW_k^\top C_kU_k^\dagger R_kV_kV_k^\top.\end{multline*}

Now let $\widetilde{R}_k:= R_kV_kV_k^\top$ and $\widetilde{C}_k = W_kW_k^\top C_k$, and notice that after some combining of terms, we have
\[(P_{T_k}(D-S_k))_{I,:} = (\widetilde{C}_k)_{I,:}U_k^\dagger R_k + (C_k-\widetilde{C}_k)_{I,:}U_k^\dagger \widetilde{R}_k,\]
and
\[(P_{T_k}(D-S_k))_{:,J} = \widetilde{C}_kU_k^\dagger(R_k-\widetilde{R}_k)_{:,J}+C_kU_k^\dagger (\widetilde{R}_k)_{:,J}.\]
Thus, the submatrices of $P_{T_k}(D-S_k)$ may be computed by keeping track of $C_k$, $U_k$, and $R_k$ from the previous iteration, and of computing only submatrices of $\widetilde{C}_k$ and $\widetilde{R}_k$ defined as above.  Thus we never form all of $P_{T_k}(D-S_k)$ at any stage, but only its constituent submatrices according to the index sets $I$ and $J$.

\section{Numerical Experiments}\label{SEC:Experiments}

Here we test the validity of the proposed algorithm with respect to other nonconvex Robust PCA solvers. Note that \citep{cai2020rapid} demonstrates that both IRCUR and AccAltProj are significantly faster and more accurate than AltProj and the gradient descent method of \citep{yi2016fast}. Thus, we only compare our algorithm to IRCUR and AccAltProj.

\subsection{Synthetic data}

\begin{figure}[h!]
    \centering
    \includegraphics[width=.45\textwidth]{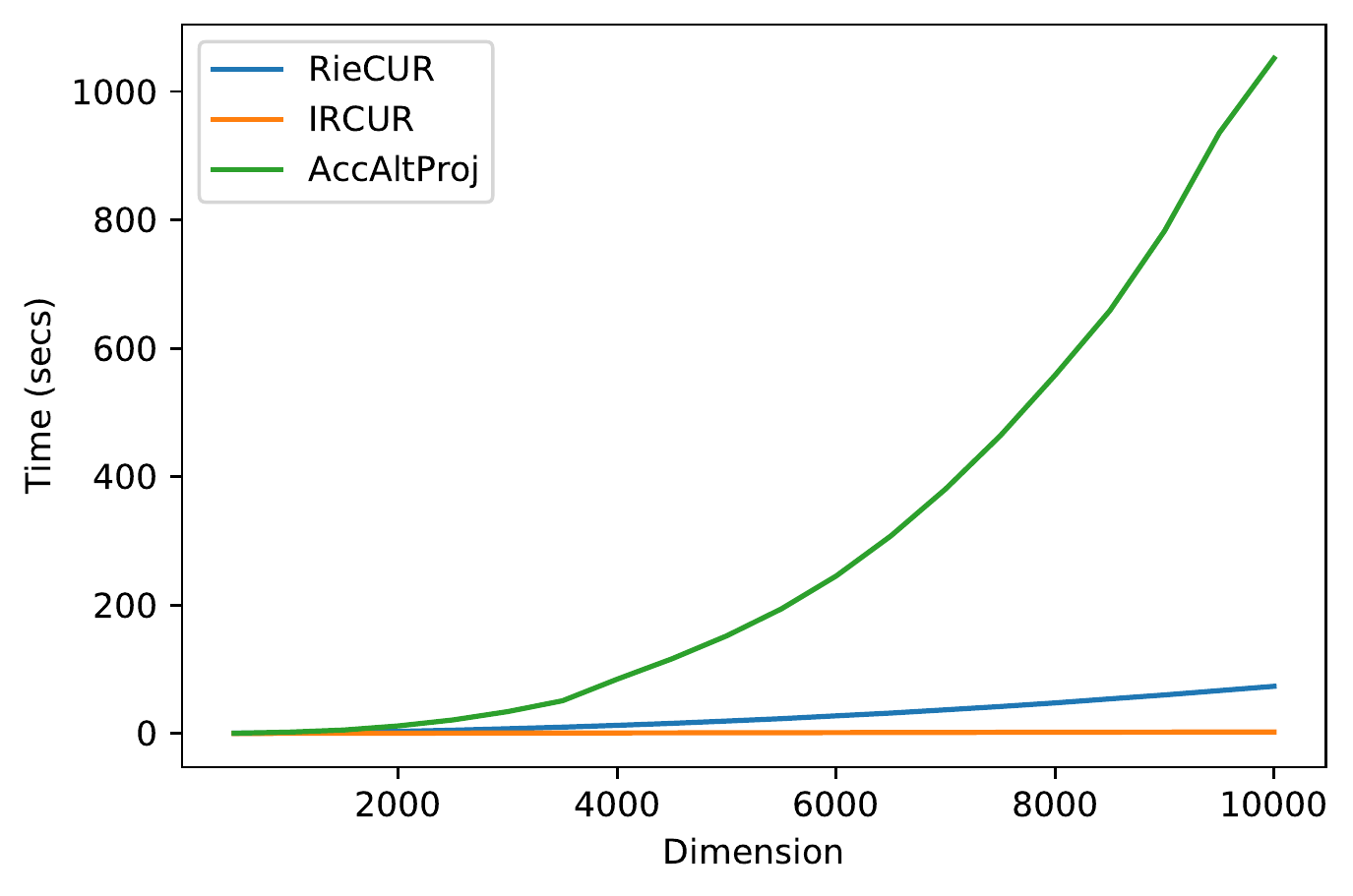}
    \caption{Time vs. Problem dimension ($n$) for the three algorithms considered. In all trials $L$ has rank $5$ and $\alpha=0.3$. Each algorithm stops once $e_k<10^{-6}$ or 40 iterations has been reached.}
    \label{FIG:TimeVDimension}
\end{figure}

First, Figure~\ref{FIG:TimeVDimension} illustrates that indeed RieCUR is significantly faster than AccAltProj, while being modestly slower than IRCUR. For each dimension $n\in\{500,1000,\dots,10000\}$, we generate a random Robust PCA problems by first generating a random low-rank matrix $L$ with rank $r=5$ as the product of two Gaussian random matrices, i.e., $L = AB^\top$ where $A,B\in\R^{n\times 5}$ have i.i.d. $\mathcal{N}(0,1)$ entries.  We generate a sparse random matrix $S$ with $\alpha = 0.3$. We set a convergence threshold of $10^{-6}$ for all algorithms, and a maximum number of iterations of 40. Figure~\ref{FIG:TimeVDimension} corresponds to average runtime over 5 random trials for each algorithm for each dimension. In all trials below, we sample $3\log n$ columns and rows when doing a CUR decomposition in RieCUR or IRCUR.

\begin{figure}[h!]
    \centering
    \includegraphics[width=.45\textwidth]{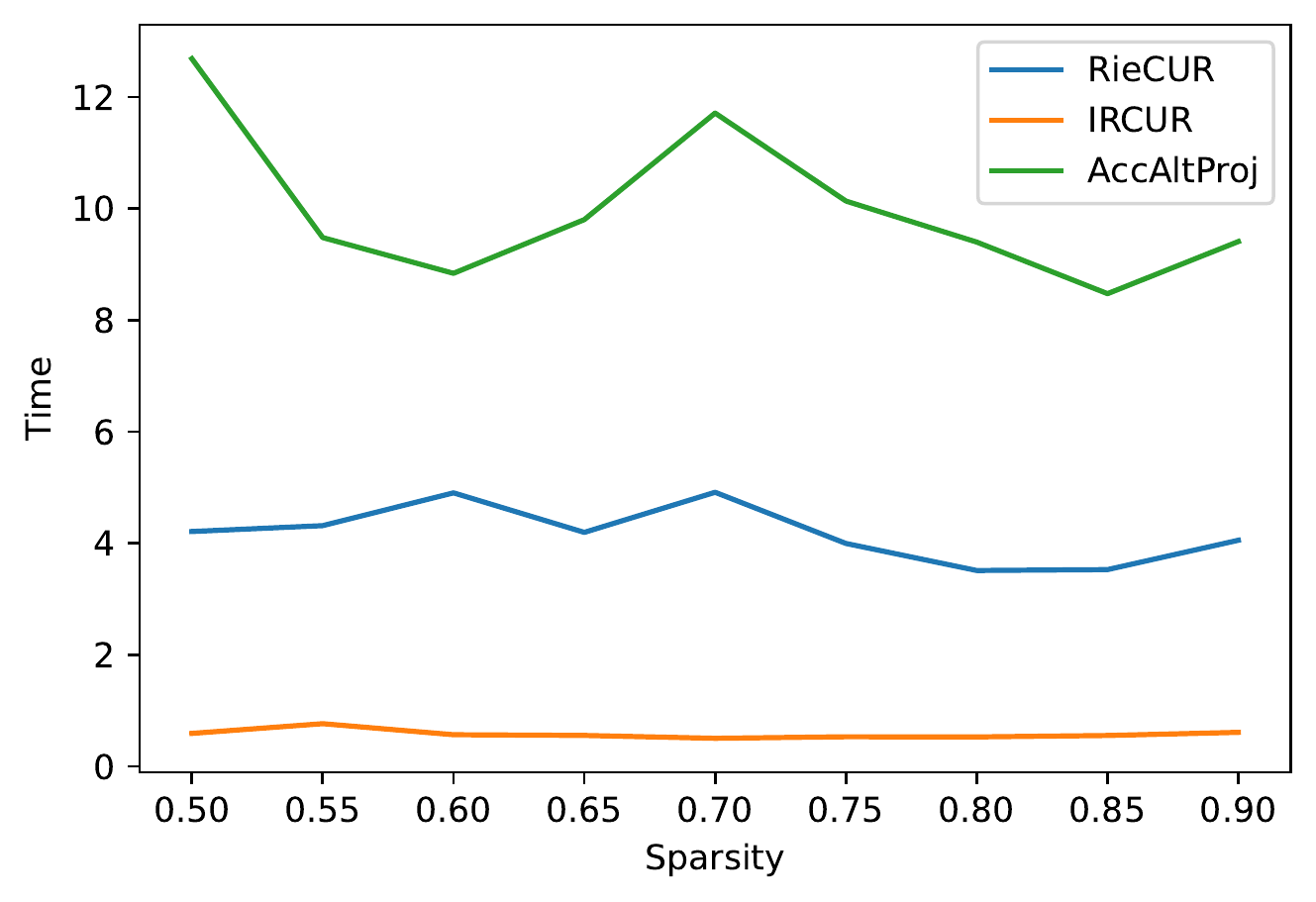}
    \caption{Time vs. Sparsity for the three algorithms considered. In all trials $L$ is $2000\times 2000$ with rank $5$. Each algorithm stops once $e_k<10^{-3}$ or 100
    iterations.}
    \label{FIG:TimeVSparsity}
\end{figure}

As a second test, we consider the effect of the sparsity of $S$ on the runtime of the algorithms. To do so, we again generate random matrices of size $2000\times 2000$, but allow $\alpha$ to range from $0.5$ to $0.95$. Each algorithm is allowed to run until either $e_k<10^{-3}$ or 100 iterations is reached, whichever comes first. The results are shown in Figure~\ref{FIG:TimeVSparsity}. One can see that RieCUR, while slower than IRCUR for all sparsity levels, remains significantly faster than AccAltProj.

\begin{figure}[h!]
    \centering
    \includegraphics[width=.45\textwidth]{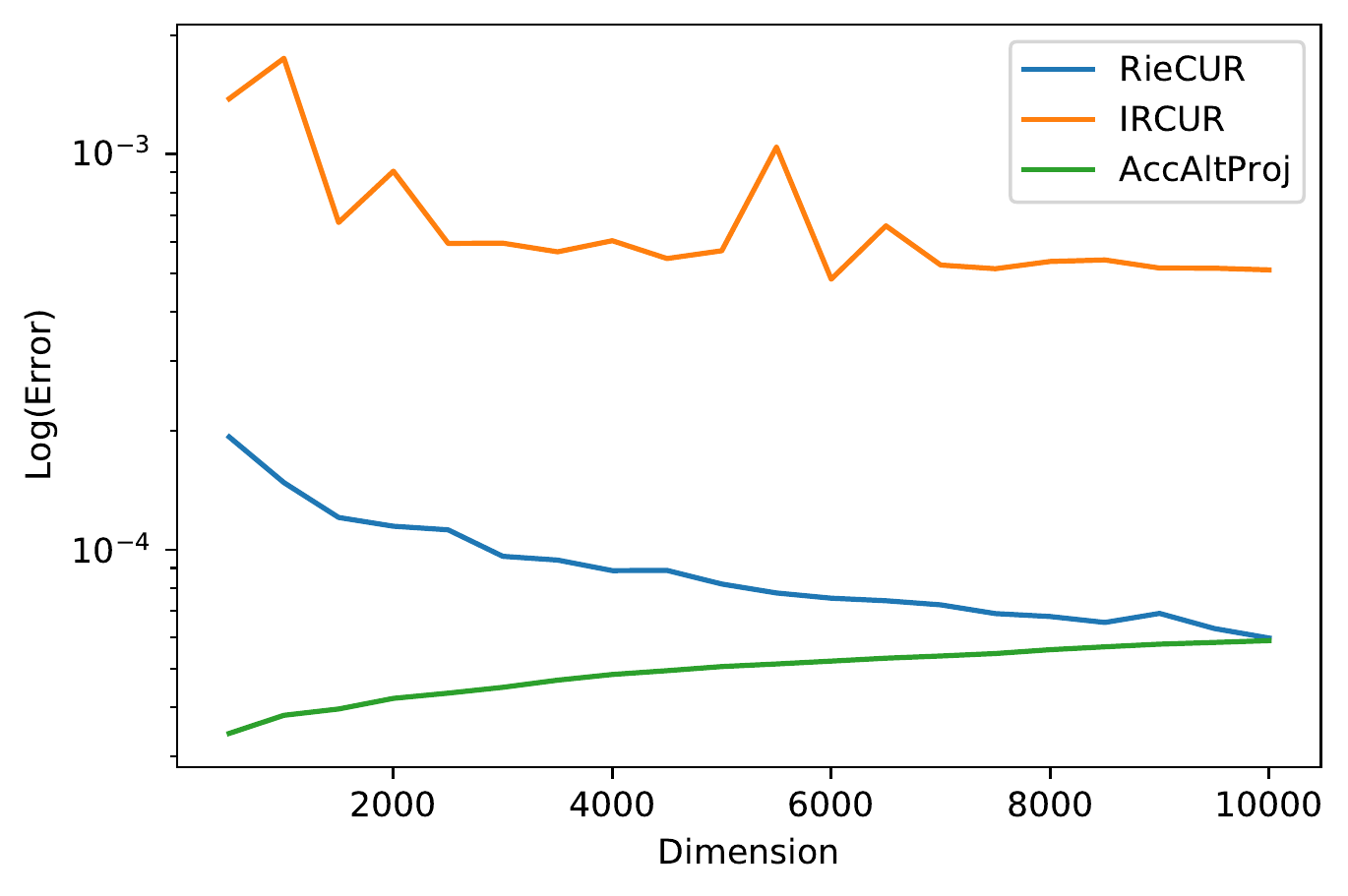}
    \caption{$\log($Final Relative Error$)$ vs. Problem dimension ($n$) for the three algorithms. In all trials, $L$ is a $2500\times2500$ rank $5$ matrix. Each algorithm stops once $e_k<10^{-6}$ or 40 iterations.}
    \label{FIG:ErrorVDimension}
\end{figure}

Next, we consider the effect of the problem dimension $n$ on the final relative error.  Figure~\ref{FIG:ErrorVDimension} shows the results for a similar setup to the above, with $L$ being a rank $5$, $2500\times2500$ matrix, and the dimension again goes from $500$ to $10,000.$ Each algorithm is set to a tolerance of $10^{-6}$ and maximum iteration of 40. One can see that IRCUR does not achieve as good error in 40 iterations compared to RieCUR which also performs slightly worse than AccAltProj.

\begin{figure}[h!]
    \centering
    \includegraphics[width=.45\textwidth]{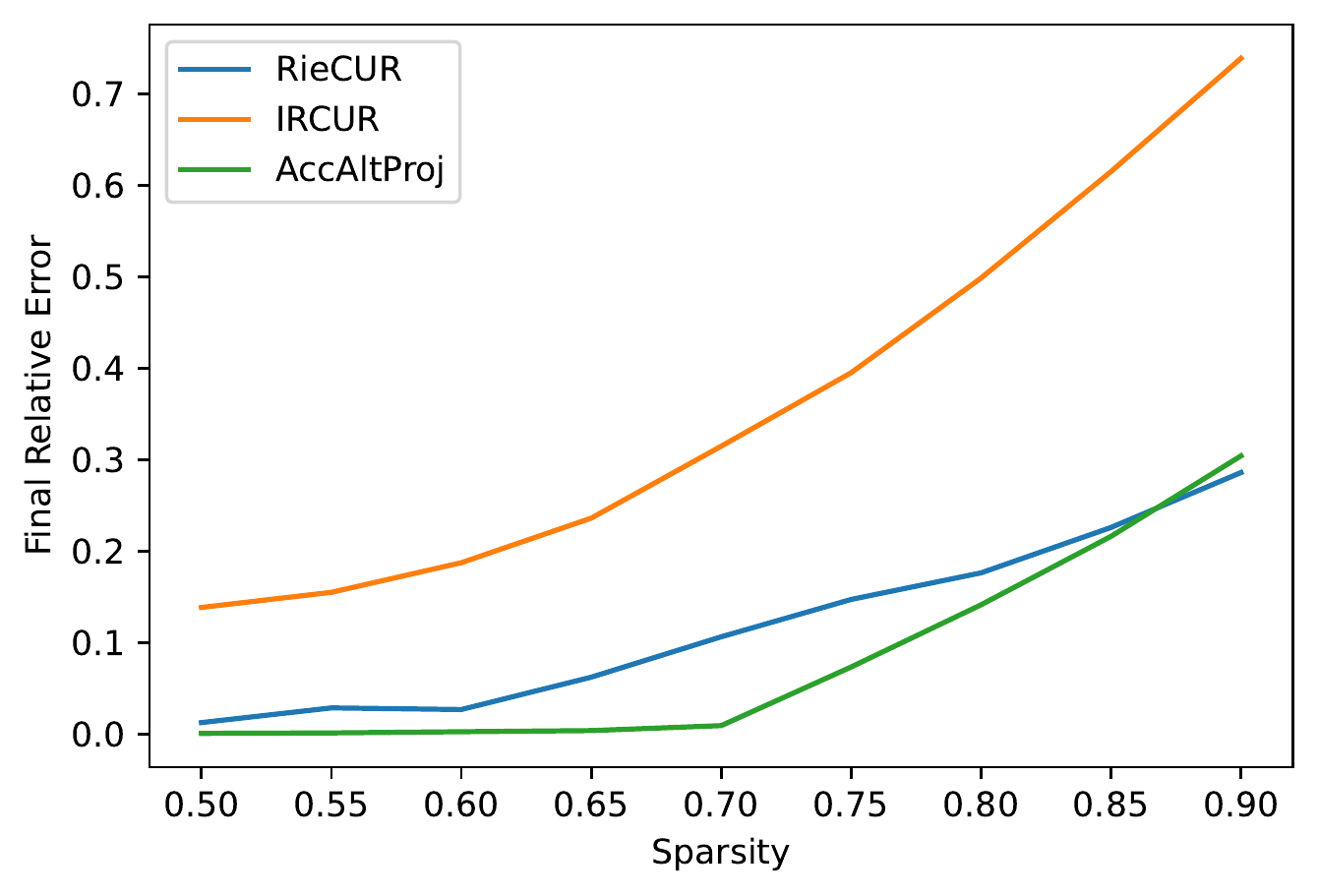}
    \caption{Error vs. Sparsity for the three algorithms considered. In all trials $L$ is $2000\times 2000$ with rank $5$. Each algorithm stops after 100 iterations has been reached.}
    \label{FIG:ErrorVSparsity}
\end{figure}

Finally, we demonstrate that RieCUR is capable of handling more outliers than IRCUR. To do so, we run all algorithms to 100 iterations and plot the final error versus the sparsity $\alpha$. We utilize the same setup as the prevous experiment for runtime versus sparsity. Note from Figure~\ref{FIG:ErrorVSparsity} that for each sparsity level, AccAltProj attains the lowest error, followed by RieCUR, followed by IRCUR. 

\subsection{Real data}

Here we illustrate the performance of RieCUR on the \texttt{restaurant} benchmark dataset.  The video is black and white with 3,055 frames of size $120\times160$. A single data matrix of the video is obtained as follows: each frame is vectorized to form a column vector of size $19,200$, and columns are concatenated into a data matrix of size $19,200\times3,055$.

\begin{figure*}[h!]
    \centering
    \includegraphics[width=.2\textwidth]{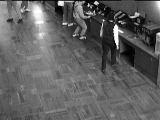}
    
    \includegraphics[width=.2\textwidth]{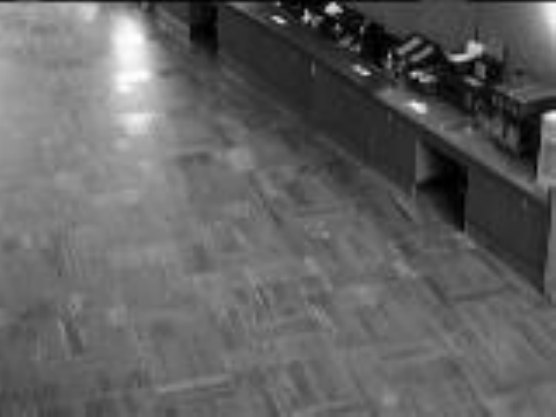}
    \includegraphics[width=.2\textwidth]{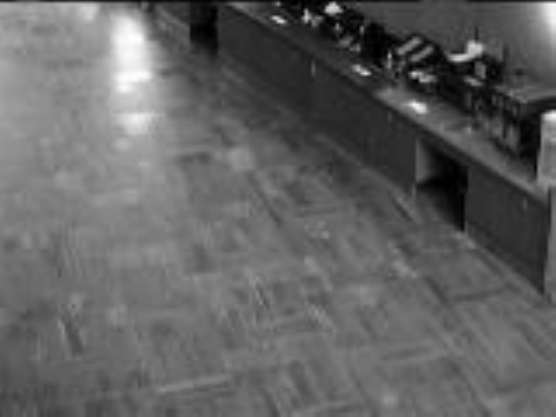}
    \includegraphics[width=.2\textwidth]{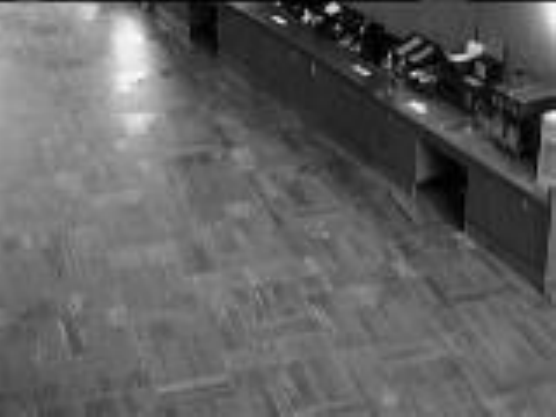}
    
    \includegraphics[width=.2\textwidth]{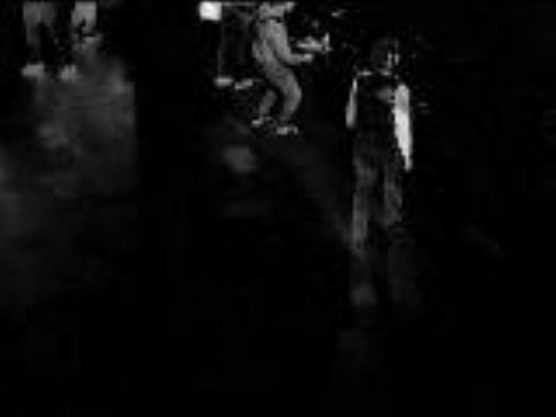}
    \includegraphics[width=.2\textwidth]{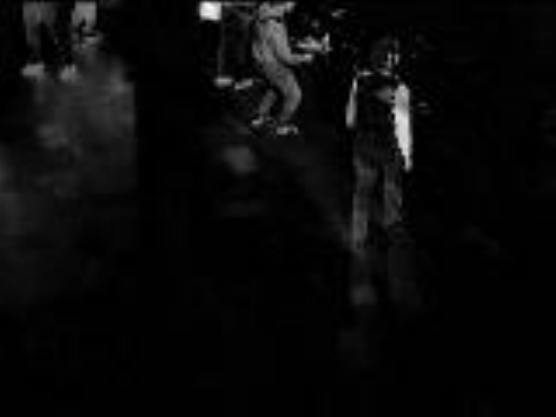}
    \includegraphics[width=.2\textwidth]{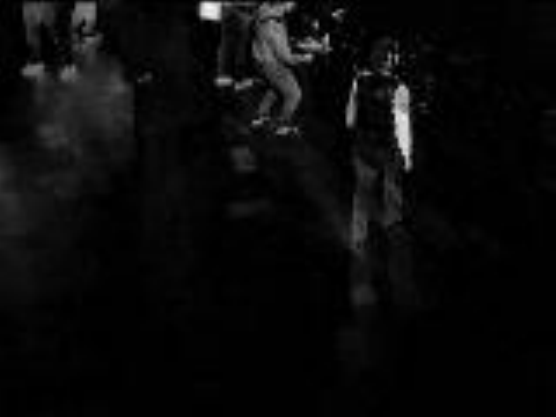}
    
    \caption{(Top row) Original frame from the \texttt{restaurant} video. (Center row) Background of the same frame recovered from AccAltProj (left), RieCUR (center), and IRCUR (right). (Bottom row) Foreground of the same frame recovered from AccAltProj (left), RieCUR (center), and IRCUR (right).}\label{fig:restaurant}
\end{figure*}

In this case, Robust PCA will separate the low-rank part of the collection of video frames, which corresponds to the static background of the video from the sparse outlier part corresponding to the foreground of the video (in this case moving people). In Figure \ref{fig:restaurant}, we show the effect of RieCUR for foreground/background separation. One can see that all algorithms tested yield high quality visual separation of the foreground and background of the videos.



\section{Conclusion}\label{SEC:Conclusion}

We have proposed Riemannian CUR (RieCUR), a nonconvex, Robust PCA algorithm, which combines ideas from both Accelerated Alternating Projections (AccAltProj) and Iterated Robust CUR (IRCUR). RieCUR has the best of both features of the other two methods, and is state-of-the art in terms of computational complexity, thus providing an alternative method to the others in cases where speed and robustness are both required.
The current algorithm carries with it a thresholding decay parameter, but future work will focus on providing an adaptive threshold which can guarantee convergence given initialization within a neighborhood of a local solution. Additionally, we use a fast starting method to initialize the guesses for the low-rank and sparse outlier matrices which uses a CUR decomposition of the corrupted data matrix $D$ rather than taking an SVD of it. This method was shown to still yield good convergence in synthetic trials while being significantly faster than standard SVD initialization.

Future work on RieCUR will consist of proving convergence to a minimizer of the nonconvex Robust PCA problem given good initialization, proof that Algorithm \ref{ALG:Init} can achieve good initialization for Robust PCA, use of a dynamic threshold rather than a static one in lines 11-13 of Algorithm \ref{ALG:Main}, and more comprehensive experiments on various Robust PCA problems.



\bibliography{refs}
\bibliographystyle{icml2022}



\end{document}